\documentclass[preprint,review,12pt]{elsarticle}

\usepackage{amssymb}
\usepackage{amsmath}
\usepackage{graphicx}
\usepackage[retainorgcmds]{IEEEtrantools}
\usepackage[colorlinks]{hyperref}
\usepackage{siunitx}
\usepackage{pdflscape}

\usepackage{enumitem}
\usepackage[normalem]{ulem}
\useunder{\uline}{\ul}{}
\usepackage{setspace} 
\usepackage[utf8]{inputenc}
\usepackage[linesnumbered, ruled, vlined]{algorithm2e}
\usepackage{rotating}
\usepackage{nicefrac}
\usepackage{todonotes}
\usepackage{hyperref}
\usepackage{xspace}
\usepackage{float}
\usepackage{caption}
\usepackage{subcaption}
\usepackage{rotating, lscape, longtable, tabu, booktabs, siunitx, multirow}
\usepackage[capitalise, noabbrev]{cleveref}
\usepackage{tkz-graph}
\usepackage{verbatim}
\usepackage{mathtools}

\newcommand{\ie}{i.\,e.\xspace}

\newcommand{\uset}[1]{\ifmmode\left\{\,#1\,\right\}\else\{\,#1\,\}\fi}

\journal{Information Processing Letters}

\begin{document}

\begin{frontmatter}

\title{On the statistical evaluation of algorithmic's computational experimentation with infeasible solutions}

\author[label1]{Iago A. Carvalho\corref{cor1}}
\ead{iagoac@dcc.ufmg.br}
\ead[url]{http://iagoac.github.io}

\address[label1]{Department of Computer Science, Universidade Federal de Minas Gerais, Brazil}

\cortext[cor1]{https://doi.org/10.1016/j.ipl.2018.11.003}

\begin{abstract}
The experimental evaluation of algorithms results in a large set of data which generally do not follow a normal distribution or are not heteroscedastic. Besides, some of its entries may be missing, due to the inability of an algorithm to find a feasible solution until a time limit is met. Those characteristics restrict the statistical evaluation of computational experiments. This work proposes a bi-objective lexicographical ranking scheme to evaluate datasets with such characteristics. The output ranking can be used as input to any desired statistical test. We used the proposed ranking scheme to assess the results obtained by the Iterative Rounding heuristic (IR). A Friedman's test and a subsequent post-hoc test carried out on the ranked data demonstrated that IR performed significantly better than the Feasibility Pump heuristic when solving 152 benchmark problems of Nonconvex Mixed-Integer Nonlinear Problems. However, is also showed that the RECIPE heuristic was significantly better than IR when solving the same benchmark problems.
\end{abstract}


\end{frontmatter}


\section{Introduction} \label{sec:intro}

Algorithms are everywhere. They perform the most varied tasks, from solving theoretical problems, such as computing the minimum spanning tree of a given graph under some constraints~\cite{Zhang2012}, to practical problems, such as routing a vehicle between two cities while minimizing the fuel cost~\cite{Papadopoulos2018}. One can compare algorithms two ways: $(i)$ theoretically; and $(ii)$ empirically. The former analyzes their runtime and space complexities, while the latter usually rely on computational experiments performed on benchmark problems.

The empirical evaluation of algorithms through computational experiment results in a large set of data that needs to be assessed by a statistical test. However, some entries of this dataset may be missing, which restricts the statistical evaluation. One cannot perform a variance test since its impossible to measure the average and the standard deviation of the dataset~\cite{Box2005}. In addition, data from computational experiments generally do not follow a normal distribution or are not heteroscedastic.~\cite{Garcia2009,Garcia2009b}. Therefore, one should use a non-parametric statistical test instead of a parametric one~\cite{Weaver2017}.

The most common approach in the literature to deal with this problem is to ignore the missing data. Although being commonly used, this strategy erroneously neglects information regarding the algorithm's behavior. The work of Pavlikov~\cite{Pavlikov2018} is an example of such a wrong practice. In the result's discussion, the author omitted the data from benchmarks where one of their algorithms cannot find feasible solutions in less than 50.000 seconds. Melo, Fampa, and Raupp~\cite{Melo2018} developed another work that has a similar approach. In their work, the benchmarks with infeasible solutions were ignored and the average result of each algorithm was reported. The work of Fortz, Oliveira, and Requejo~\cite{Fortz2017} also performed similarly and did not summarize the data from benchmarks on which an algorithm cannot find feasible solutions within their time limit. It is worth noticing that none of these works developed a statistical analysis on their data and only summarized the average behavior of their algorithms.

Hence, the objective of this work is to present a bi-objective lexicographical ranking scheme to preprocess datasets whose some of its entries are missing. The proposed scheme receives two data as input: $(i)$ the algorithms result; and $(ii)$ their running times. It analyzes the input data, ranking the algorithms for each benchmark problems. This ranking can be used as input to any statistical test. A case study is shown, using the proposed bi-objective lexicographical ranking to evaluate the experimental data of the Iterative Rounding Heuristic~\cite{Nannicini2012}. 

\section{Related works} \label{sec:related}

\sloppy
There are some alternatives to excluding the missing data. One of the most used approaches it to measure the algorithm's \emph{PAR10} score (see e.g.,~\cite{Kerschke2017,Bischl2016}). It is computed as the average runtimes of the solved benchmark problems plus ten times the cut-off time of the unsolved benchmark problems. Despite being simple and easy-to-use, it generates a large number of outliers, which can negatively affect the statistical analysis.

\fussy
The Expected Runtime Analysis (ERT)~\cite{Hansen2010,Price1997} is a well-known approach to evaluate algorithms with missing data. It measures the expected number of function evaluations
to reach a target function value for the first time. The ERT of an algorithm is computed as 
$$
    ERT = RT_S + \frac{1 - p_S}{p_S}RT_{US},
$$
where $RT_S$ and $RT_{US}$ respectively denote the average number of function for successful and unsuccessful trials, while $p_S$ denotes the proportion of successfully trials, i.e., $p_S = \frac{RT_S}{RT_{US}}$. Despite being a consolidated approach for dealing with the missing data, the ERT only estimates the algorithm's running time, which may not represent its real behavior.

The work of Campos and Benavoli~\cite{Campos2017} employed a similar strategy of the one proposed in this work. The authors proposed two statistical tests that aims at establishing a relation of dominance between algorithms by analyzing multiple performance criteria (e.g., accuracy and time complexity). The tests try to infer if there is a dominance statement that is significantly more likely than others. de Campos and Benavoli work differs from our's by manner the data is ranked. Our's methodology tries to infer if an algorithm is significantly better than other by performing a conjoint analysis of their characteristics, while de Campos and Benavoli's tests try to infer which characteristics of the algorithms statistically differ among them. 



\section{The bi-objective lexicographical ranking scheme} \label{sec:ranking}

This paper proposes a bi-objective lexicographical ranking scheme for preprocessing datasets whose some of its entries are missing. It receives two bi-dimensional vectors as input. The first vector is denoted as $\textbf{R}_{m,n}$ and contains the algorithm's results, \ie the primal solution value given by a deterministic algorithm or the average primal solution value for multiple runs of a stochastic algorithm. The second vector is denoted as $\textbf{T}_{m,n}$ and contains the algorithm's running times. As the first vector, it contains the algorithm running time, in the case of a deterministic algorithm, or the average running time of multiple runs for a stochastic algorithm. The proposed preprocessing scheme outputs a bi-dimensional vector $\textbf{A}_{m,n}$, where each column symbolizes an algorithm and each row represents a benchmark. Furthermore, each cell $a_{i,j} \in \textbf{A}$ contains an integer value that corresponds to the ranking of algorithm $j$ for the benchmark $i$.

Let $\textbf{X}_{\textbf{k}} = \{x_1, \ldots, x_m\}$ be the $k$-th line of vector $\textbf{R}$ and let $\textbf{Y}_{\textbf{k}} = \{y_1, \ldots, y_m\}$ be the $k$-th line of vector $\textbf{T}$. 
Assuming a minimization problem, one can build the $k$-th line of the matrix $\textbf{A}$ by ranking the input data according the following rules.

\begin{enumerate}
    \item If $x_i < x_j$, then $a_{ik} < a_{jk}$.
    \item If $x_i = x_j$, then $\left\{\begin{matrix}
a_{ik} < a_{jk}, & \text{if  } y_i < y_j.\\ 
a_{ik} > a_{jk}, & \text{if  } y_i > y_j.\\ 
a_{ik} = a_{jk}, & \text{if  } y_i = y_j.
\end{matrix}\right.$
    \item If $x_i$ is missing, then $\left\{\begin{matrix}
a_{ik} > a_{jk}, & \text{if  } a_{jk} \text{ is not missing}.\\ 
a_{ik} = a_{jk}, & \text{if  } a_{jk} \text{ is missing}.
\end{matrix}\right.$
\end{enumerate}

Rule $(1)$ states that, if algorithm $i$ achieves better results than algorithm $j$, then it ranked as superior. Rule $(2)$ is applied when two algorithms $i$ and $j$ achieve the same results. Then, the ranking decision is made by analyzing their running times. If algorithm $i$ runs in a smaller time than algorithm $j$, then algorithm $i$ is ranked as superior. On the other hand, if algorithm $i$ needs a greater computational time than algorithm $j$, then algorithm $j$ is ranked as superior. However, it both algorithm runs within the same time span, then they are ranked as equals. In this case, they receive the average of the ranks that would have been assigned without ties. Rule $(3)$ states that, if data from algorithm $i$ is missing (due to its inability to find feasible solutions in due time), then it is ranked equal as all other algorithms that also misses the data from the same trial. Besides, algorithm $i$ is ranked as inferior to any algorithm that finds feasible solutions in due time.

\section{A case study} \label{sec:caseStudy}

This section presents a case study of the proposed statistical test by analyzing the data provided by Nannicini and Belotti, on a work where they proposed the Rounding-based Heuristics (RBH)~\cite{Nannicini2012}. The RBH is a set of heuristics for Nonconvex Mixed-Integer Nonlinear Problems (MINLP). They are based on the idea of rounding the solution of a continuous nonlinear program subject to linear constraints. Such rounding is done by solving a Mixed-Integer Linear Program. Since there are a lot of missing solution values on the aforementioned work (due to the incapacity of the algorithms to find feasible solutions in due time), the proposed ranking scheme is especially suitable for preprocessing its results for a further statistical test.

\begin{figure}[t]
    \centering
    \includegraphics[width=\textwidth]{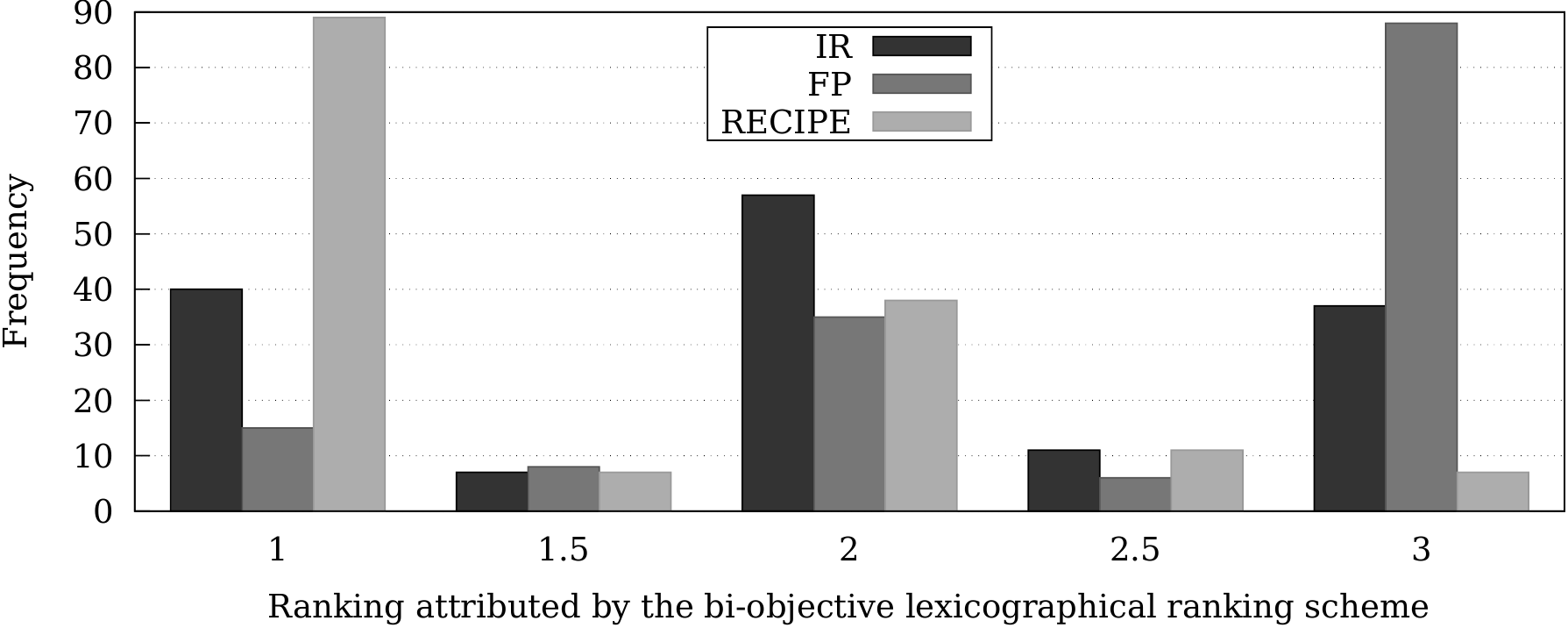}
    \caption{Histogram of the rankings attributed by the bi-objective lexicographical ranking scheme}
    \label{fig:histogram}
\end{figure}

This case study evaluated the experimental data reported in Table 9 of the above-mentioned paper. This table compared the Iterative Rounding (IR), one of the proposed RBH, with the Feasibility Pump (FP)~\cite{Ambrosio2010} and the Relaxed-Exact
Continuous-Integer Problem Exploration (RECIPE)~\cite{Liberti2011} for solving 152 difficult benchmark MINLP problems from the literature. It presents, for each instance, the primal value (vector $\textbf{R}$) and the running time (vector $\textbf{T}$) of each algorithm. 

The bi-objective lexicographical ranking scheme was then applied using the indicated vectors $\textbf{R}$ and $\textbf{T}$ as input, and a matrix $\textbf{A}$ was computed following the methodology presented in Section~\ref{sec:ranking}. Figure~\ref{fig:histogram} displays a histogram of the rankings assigned to each heuristic. According to the proposed ranking scheme, IR found better solutions than FP and RECIPE on 40 of the evaluated benchmarks. On the other hand, it was the worst algorithm on 37 out of the 152 benchmark. FP achieved the worst results among the evaluated algorithms, being inferior to IR and RECIPE on 88 benchmarks. Figure~\ref{fig:histogram} also shows that RECIPE found better solutions than IR and FP on 89 benchmarks, whereas the other algorithms outperformed RECIPE only on 7 out of the 152 evaluated benchmarks. The summation of the IR ranks was equal to 303, while the FP and RECIPE ranks summation were equal to 376 and 224, respectively. 

The Shapiro Wilk test showed that the data within matrix $\textbf{A}$ did not come from a normal distribution. For IR, it returned a $p$-value of $2.5\cdot10^{-11}$ and a $w$-statistics of $0.84$. For FP and RECIPE, the Shapiro Wilk test found $p$-values of $2.3\cdot10^{-15}$ and $2.6\cdot10^{-15}$, respectively, and a $w$-statistics of $0.73$ for both FP and RECIPE data. Therefore, the normality assumption was not verified and a non-parametric test is need.

Since this case study evaluates three algorithms, one must compare them through a multiple-comparison test, whereas the Friedman's test is the most indicated non-parametric one. The Friedman's test returned a $p$-value smaller than $2.2\cdot10^{-16}$, along with a chi-squared of $81.39$. Therefore, the null hypothesis was rejected, which means that at least one of the algorithms significantly differ from others. 

\begin{table}[!th]
\centering
\small
\caption{$p$-values and test statistics obtained by the Nemenyi's test}
\label{table:nemenyi}
\begin{tabular}{rccrcc}
\toprule
\multicolumn{3}{c}{$p$-values}    & \multicolumn{3}{c}{test statistics} \\ \cmidrule(lr){1-3} \cmidrule(lr){4-6}  
       & IR                & FP                 &        & IR     & FP      \\
FP     & $8.4\cdot10^{-5}$ & -                  & FP     & $5.92$ & -       \\
RECIPE & $1.7\cdot10^{-5}$ & $2.4\cdot10^{-14}$ & RECIPE & $6.40$ & $12.32$ \\ \bottomrule
\end{tabular}
\end{table}

In order to infer which algorithms significantly differ, this case study applied the Nemenyi–Damico–Wolfe–Dunn post-hoc test, also known as the Nemeyi's test. This test was carried out with a significance level $\alpha = 0.05$. The results of the Nemeyi's test are summarized in Table~\ref{table:nemenyi}. It shows that all algorithms significantly differ among them, since all $p$-values found are smaller than $0.05$. Therefore, one can conclude that RECIPE was the best heuristic for solving MINLP on the 152 evaluated benchmark problems, while FP was the worst one. IR demonstrated an intermediary behavior, being statistically superior to FP and statistically inferior to RECIPE.

\section*{Aknowledgments}
This study was financed in part by the Coordenação de Aperfeiçoamento de Pessoal de Nível Superior - Brasil (CAPES) - Finance Code 001.

\bibliographystyle{elsarticle-num}

\bibliography{bibsample}

\end{document}